\documentclass[lettersize,journal]{IEEEtran}
\usepackage{amsmath,amsfonts}
\usepackage{algorithm}
\usepackage{array}
\usepackage[caption=false,font=normalsize,labelfont=sf,textfont=sf]{subfig}
\usepackage{textcomp}
\usepackage{stfloats}
\usepackage{url}
\usepackage{verbatim}
\usepackage{graphicx}
\usepackage{cite}
\usepackage{nicematrix,tikz}
\usepackage{makecell}
\usepackage{hyperref}
\usepackage{placeins}

\usepackage{algpseudocode}
\usepackage{booktabs}
\usepackage{multirow}
\usepackage{pdflscape}
\usepackage{float}

\usepackage[margin=1in]{geometry}
\DeclareCaptionLabelFormat{andtable}{#1~#2  \&  \tablename~\thetable}
\DeclareCaptionLabelFormat{twotable}{#1~#2  \&  \tablename~\the\numexpr\value{table}+1}
\NiceMatrixOptions
  {
    custom-line = 
     {
       letter = I , 
       command = boldhline , 
       total-width = 1 pt , 
       tikz = { line width = 1 pt } 
     }
  }

\begin{document} 
\title{Analyzing Internal Activity and Robustness of SNNs Across Neuron Parameter Space}

\author{
    Szymon Mazurek\IEEEauthorrefmark{1},
    Jakub Caputa\IEEEauthorrefmark{1},
    Maciej Wielgosz\IEEEauthorrefmark{1}\IEEEauthorrefmark{2}

\IEEEauthorblockA{\IEEEauthorrefmark{1}Department of Computer Science, Electronics and Telecommunications, AGH University of Krakow \\
Email: {szmazurek, jakubcaputa, wielgosz}@agh.edu.pl, \IEEEauthorrefmark{2}Corresponding author}
}

\maketitle

\begin{abstract}
Spiking Neural Networks (SNNs) offer energy-efficient and biologically plausible alternatives to traditional artificial neural networks, but their performance critically depends on the tuning of neuron model parameters. In this work, we identify and characterize the existence of an \textit{operational space}—a constrained manifold in the neuron hyperparameter space (specifically membrane time constant $\tau$ and voltage threshold $v_{\text{th}}$)—within which the network exhibits meaningful activity and functional behavior. Operating inside this manifold yields optimal trade-offs between classification accuracy and spiking activity, while stepping outside leads to degeneration: either excessive energy consumption or complete network silence.

Through systematic exploration across multiple datasets and architectures, we visualize and quantify this manifold and identify efficient operating points. We further complement this analysis with experiments on robustness to adversarial noise, showing that SNNs exhibit heightened spike correlations and internal synchrony when pushed outside their operational manifold. These findings underscore the importance of principled hyperparameter tuning, not only to achieve high task performance, but also to maintain the network’s stability and energy efficiency. Our results provide practical guidelines for deploying robust and efficient SNNs, especially in neuromorphic computing scenarios.
\end{abstract}

\begin{IEEEkeywords}
Spiking Neural Networks, Neuromorphic Computing, Operational Manifold,  Adversarial Perturbation
\end{IEEEkeywords}

\section{Introduction}

Artificial Intelligence (AI) has experienced rapid advancement, largely driven by deep Artificial Neural Networks (ANNs), which have demonstrated superior performance across a wide range of applications, including vision, language processing, and robotics. Despite this success, ANNs remain difficult to interpret due to their highly overparameterized nature, and their training heavily relies on empirical hyperparameter tuning. 

Spiking Neural Networks (SNNs) represent the third generation of neural models, incorporating more biologically realistic behaviors by modeling neurons that communicate via discrete events or spikes. These models inherently capture temporal dynamics and offer significant potential for energy-efficient neuromorphic computing. However, the advantages of SNNs come at the cost of increased architectural and hyperparameter complexity. Specifically, neuron model parameters such as the membrane time constant ($\tau$) and firing threshold voltage ($v_{\text{th}}$) play a critical role in shaping the network’s behavior.

A key observation of our work is that SNNs possess an \textit{operational space} — a manifold within the neuron hyperparameter space — where the model exhibits proper activity and achieves meaningful performance. Within this manifold lies an \textit{operational point}, a specific configuration where the network is optimally tuned to balance spiking activity and task accuracy. Moving away from this region leads to network degeneration, manifesting as either excessive spiking (wasting energy) or complete inactivity (zero accuracy). This phenomenon is tightly linked to the dynamics of SNN models. For example, the "leak" mechanism of LIF neurons causes membrane potentials to decay over time in the absence of sufficient or appropriately timed inputs. When hyperparameters push the network outside the operational manifold, the leak causes the neurons to continuously lose accumulated potential, preventing spike generation and rendering the network functionally inactive. This inherent energy dissipation poses both a challenge and an opportunity for understanding the behavior of SNNs.

In contrast, second-generation networks like traditional ANNs use continuous activation functions without the same kind of temporal decay, allowing them to operate more robustly across a wider hyperparameter range. The absence of event-driven leakage in ANNs leads to smoother performance surfaces and alleviates the need to maintain operation within a sharply bounded manifold.

This work addresses the need to better understand and navigate the operational space of SNNs. We propose a systematic study of internal spiking activity in relation to $\tau$ and $v_{\text{th}}$, revealing how these parameters shape the behavior and energy profile of the network. Our analysis focuses on two major objectives: \textit{(1)} identifying and characterizing the manifold of operational points where SNNs function effectively, and \textit{(2)} evaluating the impact of input perturbations on internal activity patterns and robustness.

We demonstrate that the internal activity and energy consumption of an SNN are highly sensitive to its neuron parameter configuration. We further provide empirical evidence for the existence of efficient configurations—operational points—that deliver strong accuracy with minimal spike activity. Additionally, under distributional shifts such as increased input noise, we observe that SNNs exhibit heightened spiking activity and inter-neuronal correlations, revealing susceptibility to adversarial or noisy conditions.

Our findings highlight the practical importance of understanding and selecting the right operating regime in the high-dimensional hyperparameter landscape of SNNs. Beyond performance optimization, this perspective also opens avenues for diagnosing and mitigating failures due to input noise, data drift, or model misconfiguration—contributing to more robust and energy-aware neuromorphic computing systems.

\section{Related Work}

Despite the ubiquitous complexity, there are efforts to structure and formalize the understanding of the inner mechanisms governing neural networks and processes of their creation and usage. In one of the most notable research on this topic, Martin and Mahoney \cite{Charles2021WeightWatch} utilize Random Matrix Theory to study the change in ANN parameters distributions during training. They identify phases thought which layer weights move during training, as well as observe those states in trained models. Their approach allows for detailed tracking of model state beyond classical metrics.

Another important research direction in that domain is the analysis of loss landscapes. Hao et al. \cite{Hao2017LossLandscape} explore methods for visualizing the loss landscape of a network, using them to observe influence of network architecure on its ability to converge and generalize. 

Significant advances have also been made in research regarding methods for automatic hyperparameter choice in ANN design, the so-called AutoML. Extensive literature shows that importance and difficulty of hyperparameter tuning is recognized, with numerous solutions allowing to optimize the model's architecture and hyperparameter choice \cite{Salehin2024automl}. Their full analysis is beyond the scope of this article.

In the domain of SNNs, such methods from the ANN domain are also applicable, as SNN models do posses similar properties (trainable weights, hierarchical layered structure). However, to our knowledge, there are no direct attempts to create tools similar to \cite{Charles2021WeightWatch} dedicated to SNN analysis.
Additionally, one may expect that due to the increased complexity of SNN via their inherent temporal properties and larger number of hyperparameters, the extrapolation of ANN-based analysis methods and AutoML tools can become difficult. Despite that, attempts of such method transfers start to emerge, especially with AutoML methods \cite{Firmin2024ParallelAutoml, Honda2025snnautoml} . However, they are often limited by focusing primarily on end task performance, without consideration for parameter choice influence on network's internal activity, and therefore energy usage.

As mentioned previously, deep analyses of SNNs, utilized as machine learning tools, are limited. They are widely explored in neuroscience, with primary focus on recurrent neural circuits as fundamental computational units across the brain \cite{Edmund2010attractor,Khona2022Attractor}. Despite these contributions, neuroscience focuses on understanding rules governing biological systems, rather than creating robust systems capable of solving common machine learning problems, where model architectures and their properties may not strictly adhere to rules observed in biology.

In this work, we aim to address this gap by exploring insights provided by internal activity of SNN models solving concrete learning tasks, along with their dependence on initial conditions preceeding training (i.e hyperparameter choice).

\section{Methods}

\subsection{Foundational Concepts of Spiking Neural Networks}
Although ANNs have achieved remarkable performance across various domains, an attempt to leverage their established architectural successes within the SNN domain has shown considerable promise, producing competitive results \cite{FriedemannRobustnessSurrogate}. However, a direct porting of ANN designs to SNNs is complicated by fundamental differences in their operational principles, which require specific adaptations to the models \cite{Ding2021ANNSNN}. This section will formally define the core operational mechanisms governing SNNs, neuron models, and surrogate gradient method.

\subsubsection{Core Principles of Spiking Neurons}
Within the architecture of SNNs, individual neurons operate by integrating incoming signals and generating discrete output events, or "spikes," which are typically represented as binary vectors over time. This fundamentally distinguishes them from ANNs, where continuous activation functions are employed. The transformation of ANNs into SNNs primarily involves substituting these continuous activations with models that emulate the spiking behavior of biological neurons.

At each discrete timestep $t$, the internal state, or membrane potential, of a neuron, denoted as $H[t]$, is determined by its previous state and the current input. This update mechanism is expressed as:
\begin{equation}
H[t] = f(V[t-1], X[t]),
\end{equation}
where $X[t]$ signifies the input vector at time $t$, $V[t-1]$ represents the membrane potential after a potential spike at the previous timestep, and $f$ is the specific neuron update function, which varies based on the chosen neuron model and will be elaborated upon subsequently.

Upon the emission of a spike, a neuron undergoes a reset of its membrane potential. This reset mechanism can be implemented in various ways, but for this study, a "hard reset" approach is adopted. In this scheme, the membrane potential is instantaneously reduced following a spike. The membrane potential $V[t]$ at timestep $t$ is thus governed by:
\begin{equation}
V[t] = H[t] \cdot (1-S[t]) + V_{reset} \cdot S[t],
\end{equation}
Here, $S[t]$ indicates whether a spike was fired at time $t$, and $V_{reset}$ is the fixed potential to which the neuron's membrane potential is reset after firing.

The decision for a neuron to fire a spike is contingent upon its membrane potential exceeding a predefined threshold. The neuronal firing function $S[t]$ is formally defined by:
\begin{equation}
S[t] = \Theta(H[t] - v_{th}),
\end{equation}
where $v_{th}$ denotes the fixed threshold voltage. The function $\Theta$ is the Heaviside step function, mathematically expressed as:
\begin{gather}
\Theta(x) =
\begin{cases} 1, \; x\geq0 \\0, \; x<0 \end{cases}
\end{gather}
This implies that a spike ($S[t]=1$) is generated only when the membrane potential $H[t]$ reaches or surpasses the threshold $v_{th}$.

\subsubsection{The Leaky Integrate-and-Fire Neuron Model}
A spectrum of neuron models, varying in their degree of biological fidelity and computational demands, are available for SNN simulations \cite{Izhkievich2004Neurons}. For the purposes of this research, we have opted for the Leaky Integrate-and-Fire (LIF) model \cite{Lu2022Lif}. This model is favored for its optimal balance between capturing essential neuronal dynamics and maintaining computational tractability, making it a highly efficient choice.

The update rule for the membrane potential in the LIF neuron model, corresponding to the function $f$ introduced earlier, is specifically defined as:
\begin{equation}
f(V[t-1], X[t]) = V[t-1] + \frac{1}{\tau}(X[t] - (V[t-1] -V_{reset})),
\end{equation}
In this formulation, the term $\frac{1}{\tau}$ (where $\tau > 1$) represents the membrane time constant. This constant is crucial as it dictates the rate at which the neuron's membrane potential "leaks" or decays back towards its resting state in the absence of input, thereby influencing the neuron's responsiveness to incoming spikes.

\subsubsection{Surrogate Gradient Training for Spiking Neural Networks}
The inherent event-driven and sparse nature of SNNs, while advantageous for energy efficiency, presents a significant hurdle for training. Specifically, the non-differentiable nature of the Heaviside step function, which dictates spike generation, renders standard gradient-based optimization techniques (like backpropagation, widely used in ANNs) inapplicable. The "surrogate gradient" method emerges as a powerful solution to this challenge, facilitating the application of gradient descent optimization to SNNs by approximating the discontinuous firing function with a continuous, differentiable counterpart \cite{NeftciSurrogateGL}.

During the forward propagation phase, the neuron's response adheres strictly to its original definition, with spike generation determined by the discontinuous Heaviside function $\Theta(x)$. However, the mathematical derivative of this function, which is required for backpropagation, is the Dirac delta function:
\begin{gather}
\Theta'(x) =
\begin{cases} \infty, \; x=0 \\0, \; x\neq0 \end{cases}
\end{gather}
This characteristic makes direct gradient computation infeasible. To circumvent this, during the backward pass (gradient computation), the Heaviside function is replaced by a carefully chosen continuous and differentiable "surrogate" function, denoted as $g$. For this investigation, $\arctan$ and $sigmoid$ functions were utilized for distinct analytical purposes: $\arctan$ for neuron hyperparameter analysis and $sigmoid$ for examining internal activity changes under noisy conditions.

Consequently, for the purpose of gradient calculation during the backward pass, the effective firing function $S[t]$ is modified to:
\begin{equation}
S[t] = g(H[t]-v_{th}),
\end{equation}
This substitution enables the computation of meaningful gradients, thereby allowing for the effective backpropagation of errors and the optimization of network parameters.

\subsection{Analysis of network performance and spiking activity dependent on neuron model parameters}

In this experimental investigation, we explored two distinct SNN architectures: a Convolutional SNN (CNNSNN) and a Multilayer Perceptron SNN (MLPSNN). Both architectures incorporated Leaky Integrate-and-Fire (LIF) neurons, characterized by a membrane time constant $\tau$ and a voltage threshold $v_{\text{th}}$.

The MLPSNN comprised three fully connected LIF layers, designed to process flattened image inputs. To enhance the capture of spatial patterns in image data, a deeper CNNSNN was implemented. This CNNSNN architecture featured three convolutional layers followed by two linear LIF layers.

For the MLPSNN, input images were transformed into spike trains using a Poisson encoder. This encoder used pixel values as the probability of spike generation in the $T$ timesteps. Network optimization was performed using the Adam optimizer \cite{Kingma2014AdamAM} with a learning rate of $10^{-3}$. The objective was to minimize the mean squared error (MSE) loss between the network's output firing rate and the target label. The network output itself was calculated as the average firing rate over $T$ output timesteps.

This setup was employed to train both network architectures to convergence on the MNIST \cite{Deng2012mnist} and CIFAR10 \cite{Krizhevsky09learningmultiple} datasets. A grid search was conducted across various combinations of $\tau$ and $v_{\text{th}}$ values. For each combination, the resulting test accuracy and the total number of spikes emitted by the network during inference on the test dataset were recorded.

We also use the collected data to estimate the Pareto-optimal neuron parameter configuration that leads to both robust performance and low energy usage. To do so, we calculate the network's efficiency $\eta$ as:
\begin{equation}
    \eta = \frac{\text{Normalized test accuracy}}{\text{Normalized spike count}}
\end{equation}







\subsection{Adversarial noise conditions}
To assess the changes in activity pattern and adversarial robustness of SNNs against input perturbations, we devised a systematic experimental procedure involving targeted noise injection and activation analysis.

Contrary to  the previous experiments, we employed two advanced SNN architectures: SewResnet18\cite{Fang2021DeepRL} and Spiking VGG11\cite{Simonyan2014VeryDC}. Both were trained on the MNIST and CIFAR10 datasets. 
The network output was obtained similar to previous experiments.
For input encoding, we employ convolutional layers to adaptively convert pixel values into spikes. For both training and inference, samples were presented to the network with a fixed timestamp duration of $T=10$. Each input image was repeated for $T$ time steps within a given sample.

Each model was trained until convergence, optimizing the cross-entropy cost function with Adam optimizer \cite{Kingma2014AdamAM} for maximum of 100 epochs. The learning rate was set to $10^{-2}$ and the batch size to 128. Early stopping procedure was employed, halting the training if no improvement of validation loss was observed for more than 5 epochs. Validation set was extracted as 10\% of the training data. Model that provided lowest validation loss was used for further analyses. LIF remained as a spiking neuron model with sigmoid as approximating function for surrogate gradient. 

Following the training of all models to convergence on each of the datasets, we initiated a noisy sample injection procedure to evaluate their resilience to input corruption. This procedure was applied to 2000 samples randomly selected from the respective testing datasets for each model. For each selected source test sample, an initial prediction was made. If it was correct, the activation maps of every layer across $T$ time steps of the network were recorded and stored as baseline references. Subsequently, an iterative process of input perturbation was initiated. In each iteration, a randomly selected input frame (out of the $T$ frames) was replaced with normal Gaussian noise. After injecting noise into one frame, the perturbed sample was fed to the network and a new prediction was obtained. This procedure was repeated, progressively increasing the number of corrupted frames in the input sequence with each step. The process continued until the network produced an incorrect prediction for the given perturbed sample, marking the network's failure to correctly classify the increasingly noisy input. 
As for the clean sample, once again the activation matrices were collected after the network failure.
This process yielded a set of activation maps for both clean and progressively corrupted inputs for each analyzed sample.

Upon obtaining the activation maps for all predictions across the test samples, we proceeded to compute average correlation matrices. For each layer, we computed the Pearson correlation scores between the spike trains (activations) of that layer. For a given network and a particular layer, we calculated the correlation matrix for all activations obtained across each recorded sample for both clean sample and the one for which noise injection caused an incorrect prediction. These individual correlation matrices were then averaged by the total number of samples, yielding a representative average correlation matrix for that specific layer. 
The experimental procedure is formally described in Algorithm \ref{alg:noisy_injection}.

\begin{algorithm}
\caption{Procedure for analyzing spike pattern changes in SNN under noisy input conditions}
\label{alg:noisy_injection}
\small
\begin{algorithmic}[1]
\Require Trained SNN model $M$, test dataset $D$, input time steps $T=10$, sample size $N=2000$
\Ensure Average correlation matrices for each layer

\State Initialize empty correlation matrix storages $C^{clean}, C^{cor} $
\State Randomly select $N$ samples from test dataset: $S = \{s_1, s_2, \ldots, s_N\} \subset D$

\For{each sample $s_i \in S$}
    \State Create frames sequence: $x_i = [s_1, s_2, \ldots, s_T]$
    
    \If {$\hat{y}_i^{clean} = M(x_i)$ incorrect}
        \State Skip further analysis
    \EndIf
    \vspace{0.15cm}
    \State \textbf{// Collect spike train matrices (activation) for each layer after correct prediction on clean sample}
    \For{each layer $l \in \{1, 2, \ldots, L\}$}
        \State $A_{l}^{clean} = \{a_{1}^{clean}, a_2^{clean}, \ldots, a_L^{clean}\}$
    \EndFor
    \vspace{0.15cm}
    \State \textbf{// Start noise injection procedure}
    \State Initialize corrupted input: $x_i^{cor} = x_i$
    \State Initialize list of corrupted frame indices: $T^{cor} = \{\}$
    \While{$\hat{y}_i^{cor}$ correct}
        \State $t^{cor} \gets \text{SAMPLE}(T \notin T^{\text{cor}})$
        \State $T^{cor}.APPEND(t^{cor})$
        \State $x_i^{cor}[t^{cor}] = \mathcal{N}(0, 1)$
        \State $\hat{y}_i^{cor} = M(x_i^{cor})$
    \EndWhile
    \vspace{0.15cm}
    \State \textbf{// Collect spike train matrices (activation) for each layer after incorrect prediction on noisy sample}
    \For{each layer $l \in \{1, 2, \ldots, L\}$}
        \State $A_{l}^{cor} = \{a_{1}^{cor}, a_2^{cor}, \ldots, a_L^{cor}\}$
    \EndFor

\EndFor
\vspace{0.15cm}

\State \textbf{// Compute average spike train Pearson correlations across all recorded samples for clean and corrupted ones for each layer}
\For{each layer $l \in \{1, 2, \ldots, L\}$}
    \State $C_{l}^{clean} = \frac{1}{N} \sum_{i=1}^{N} corr(a_{i,l}^{clean})$
    \vspace{0.15cm}
    \State $C_{l}^{cor} = \frac{1}{N} \sum_{i=1}^{N} corr(a_{i,l}^{cor})$
    
\EndFor

\State \Return $C^{clean}, C^{cor}$
\end{algorithmic}
\end{algorithm}

This ratio balances the trade-off between accuracy and spiking activity, favoring parameter settings that achieve high accuracy with fewer spikes. It serves as an indicator of energy-efficient performance.

\section{Results}

\subsection{Neuron parameter's influence on SNN performance and internal activity}
\begin{figure*}[hb!]
    \centering

    \subfloat[CIFAR CNNSNN]{
        \includegraphics[width=0.48\textwidth]{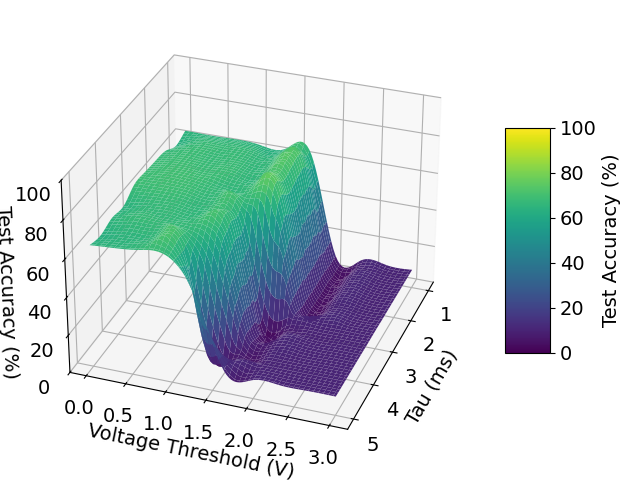}
    }
    \hfill
    \subfloat[CIFAR MLPSNN]{
        \includegraphics[width=0.48\textwidth]{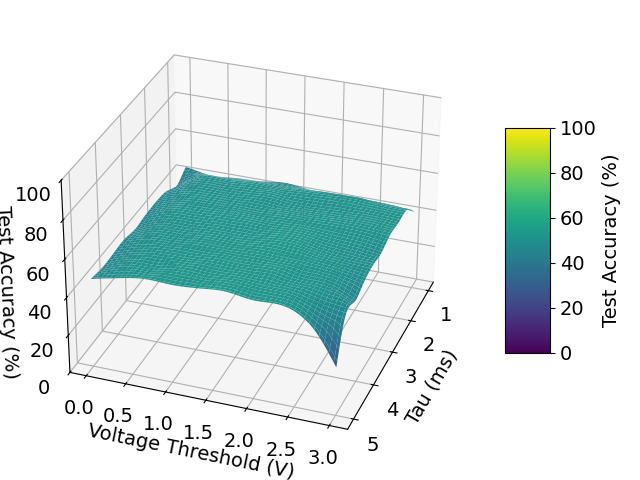}
    }

    \vspace{0.3cm}

    \subfloat[MNIST CNNSNN]{
        \includegraphics[width=0.48\textwidth]{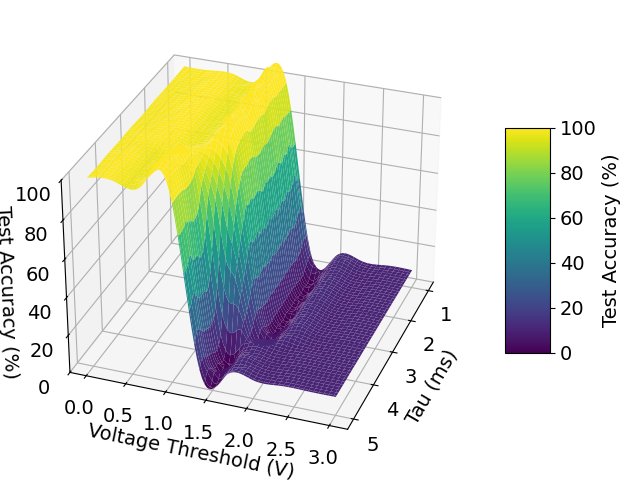}
    }
    \hfill
    \subfloat[MNIST MLPSNN]{
        \includegraphics[width=0.48\textwidth]{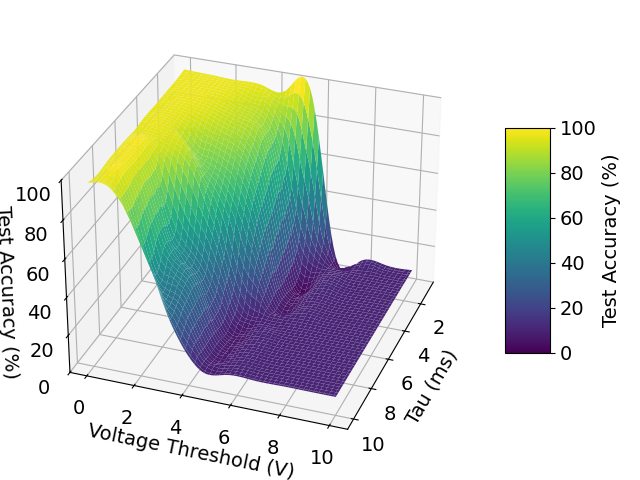}
    }

    \caption{Test accuracy of models trained on CIFAR and MNIST as a function of neuron membrane threshold and decay constant $\tau$.}
    \label{fig:test-acc}
\end{figure*}

\begin{figure*}[hb!]
    \centering
    \subfloat[CIFAR CNNSNN]{
        \includegraphics[width=0.48\textwidth]{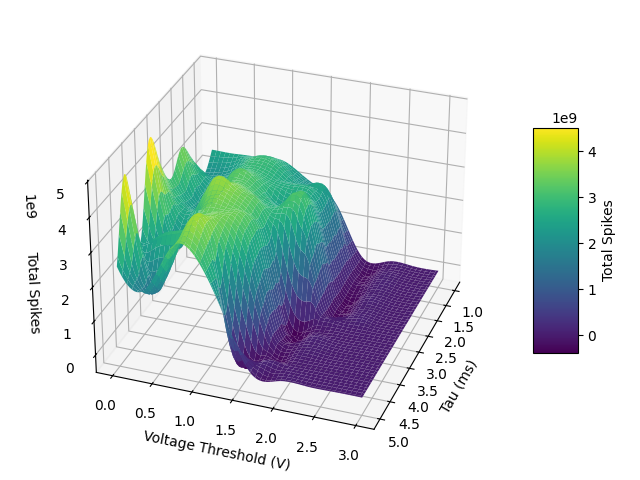}
    }
    \hfill
    \subfloat[CIFAR MLPSNN]{
        \includegraphics[width=0.48\textwidth]{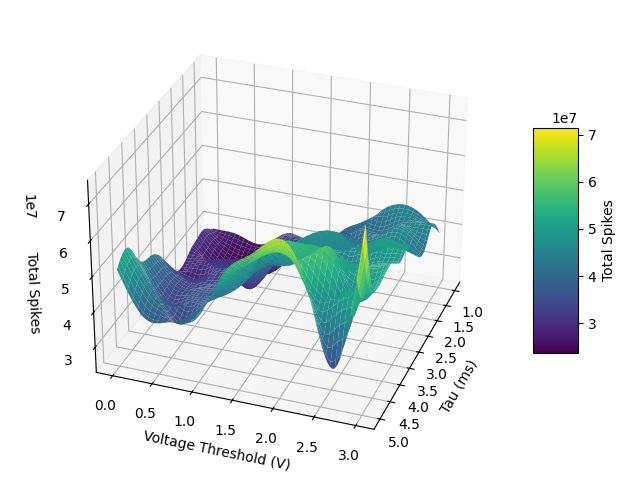}
    }

    \vspace{0.3cm}

    \subfloat[MNIST CNNSNN]{
        \includegraphics[width=0.48\textwidth]{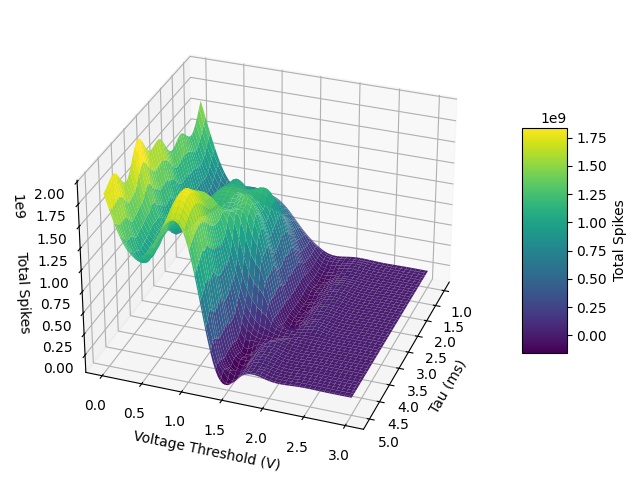}
    }
    \hfill
    \subfloat[MNIST MLPSNN]{
        \includegraphics[width=0.48\textwidth]{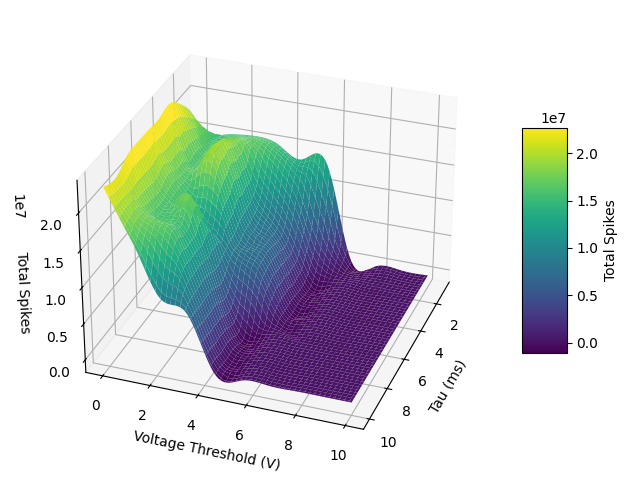}
    }

    \caption{Total number of spikes fired by the network during inference using trained models on CIFAR10 and MNIST datasets as a function of neuron membrane threshold and decay constant $\tau$.}
    \label{fig:spike-count}
\end{figure*}

\begin{table*}[hb]
\centering
\caption{Performance and spiking activity of most efficient models, with $\tau$ and $v_{th}$ hyperparameters, for each dataset.}
\label{tab:optimal_points}
\begin{tabular}{ccccccc}
\hline
\textbf{Dataset} & \textbf{Architecture} & \(\tau\) & \(V_{\text{th}}\) & \textbf{Test Acc. (\%)} & \textbf{Total Spikes} & \textbf{Efficiency} \\
\hline
MNIST   & CNNSNN & 1.44 & 0.67 & 76.23 & \(2.50 \times 10^7\) & \textbf{430.88} \\

MNIST   & MLPSNN  & 1.001  & 1.33 & 99.01 & \(2.54 \times 10^8\) & 7.74 \\

CIFAR10 & CNNSNN & 1.001 & 1.67 & 63.07 & \(1.44 \times 10^9\) & 2.95 \\

CIFAR10 & MLPSNN  & 3.00 & 4.45 & 49.00 & \(6.77 \times 10^6\) & 2.62 \\
\hline
\end{tabular}
\end{table*}

\begin{table*}[ht]
\centering
\caption{Performance and spiking activity of models with highest accuracy, with $\tau$ and $v_{th}$ hyperparameters, for each dataset.}
\begin{tabular}{cccccc}
\toprule
\textbf{Dataset} & \textbf{Architecture} & $\tau$ & $V_{\text{th}}$ & \textbf{Test Acc. (\%)} & \textbf{Total Spikes} \\
\midrule
MNIST & CNNSNN & 1.88 & 1.3389 & 99.1 & $7.10 \times 10^8$ \\
MNIST & MLPSNN & 1.001 & 0.01 & 97.02 & $1.29 \times 10^7$ \\
CIFAR10 & CNNSNN & 2.77 & 0.3422 & 69.71 & $2.27 \times 10^9$ \\
CIFAR10 & MLPSNN & 5.00 & 1.6711 & 53.23 & $6.18 \times 10^7$ \\
\bottomrule
\label{tab:top_configs_basic}
\end{tabular}
\end{table*}

Exploration of the membrane time constant ($\tau$) and voltage threshold ($v_{th}$) reveals an existing relationship between these neuron hyperparameters, performance, and internal spiking activity in tested networks. It can be seen in Figures \ref{fig:test-acc} and \ref{fig:spike-count}. 

On MNIST dataset, both the CNNSNN and MLPSNN architectures demonstrate strong performance. The CNNSNN achieves near-perfect test accuracy, peaking at 99.36\%, which is slightly higher than the MLPSNN's best of 97.57\%. We observe that the highest accuracy for the CNNSNN is typically achieved at a moderate voltage threshold, approximately $\sim1.34$ to $\sim1.67$ V. In contrast, the MLPSNN's performance degrades more sharply with increasing voltage thresholds compared to the CNNSNN. 

On the more complex CIFAR-10 dataset, the overall test accuracy for both models is, as anticipated, lower than on MNIST. Here, the CNNSNN significantly outperforms the MLPSNN, achieving a best test accuracy of 69.46\% versus 53.53\%. Both models generally achieve their best results at lower to mid-range voltage thresholds, specifically around $\sim0.34$ V for CNNSNN and $\sim1.67$ V for MLPSNN. 

Analysis of total spike counts reveals distinct behaviors across architectures and datasets. CNNSNN models consistently exhibit higher spike counts than MLPSNN due to their more complex architecture. A critical observation is that spike activity sharply decreases when voltage thresholds exceed a certain point.
This reduction in spiking often correlates with the network's test accuracy dropping to chance level, indicating an effective cutoff in network activity.

It was also noted that CNNSNNs demonstrate greater resilience to increases in voltage threshold, maintaining high accuracy up to moderate levels. In contrast, MLPSNNs are more sensitive to elevated thresholds, where high values lead to rapid degradation of both spike count and accuracy. 

As summarized in Tables \ref{tab:optimal_points} and \ref{tab:top_configs_basic}, a distinct trade-off exists between classification accuracy and computational cost, measured by total spike activity. For instance, the MNIST CNNSNN model, while achieving an impressive 99.36\% test accuracy, incurs a cost of over $\sim7.1 \times 10^8$ spikes (Table \ref{tab:top_configs_basic}). However, a more efficient configuration can reduce the spike count by an order of magnitude to approximately $\sim2.5 \times 10^7$ lowering the accuracy in exchange (76.23\%, Table \ref{tab:optimal_points}). Similarly, in the CIFAR10 MLPSNN case, an efficiency-focused tuning (Table \ref{tab:optimal_points}) results in only about $\sim6.8 \times 10^6$ spikes, but accuracy decreases to 49.00\%, which is notably lower than the 53.53\% achieved by an accuracy-optimized setup that consumes nearly an order of magnitude more resources.

CNNSNN models consistently outperform MLPSNNs in test accuracy, particularly on intricate datasets like CIFAR10, though this comes with a higher spike cost. MLPSNNs tend to operate more efficiently (fewer spikes) but reach lower performance ceilings. These discrepancies may also originate from differences in network sizes (trainable parameter counts). 
The interaction between voltage threshold and membrane time constant is non-linear. While lower thresholds facilitate dense spiking and high accuracy, increasing the threshold drastically reduces spike count, often at the cost of accuracy. The optimal values for these parameters vary significantly across different architectures and datasets, emphasizing the necessity of dataset-specific tuning. However, comparing Figures \ref{fig:test-acc} and \ref{fig:spike-count}, it can be seen that there exist hyperparameter configurations that can decrease the network's internal activity while maintaining high performance.

Overall, it can be noted that neuron hyperparemeters can be tuned to achieve different performance and energy usage levels, a consideration particularly vital for deployment on neuromorphic hardware. The choice of operating point should align with specific application constraints: tasks where accuracy is paramount should favor settings similar to those in Table~\ref{tab:top_configs_basic}, while energy-constrained setups prioritizing spike efficiency are better served by configurations similar to those in Table~\ref{tab:optimal_points}.

\subsection{Analysis of layer spike activity under input perturbation conditions}

Initially, we trained the networks for further evaluation under noisy input. All networks have converged, achieving relatively high performance. The results are shown in Table \ref{tab:test-acc-adv}. The networks perform similarly on a given dataset, with expected higher performance on MNIST dataset. As achieving maximum performance was not our goal in these experiments, we proceeded with the analyses using those models.

\begin{table}[ht]
\centering
\caption{Test accuracies achieved by the networks used for evaluating noise perturbation effects}
\begin{tabular}{ccc}
\toprule
\textbf{Dataset} & \textbf{Architecture} & \textbf{Test Acc. (\%)}  \\
\midrule
MNIST & SewResnet18 &  99.04 \\
MNIST & Spiking VGG11  & 98.65 \\
CIFAR10 & SewResnet18 & 69.39  \\
CIFAR10 & Spiking VGG11 & 70.86  \\
\bottomrule
\label{tab:test-acc-adv}
\end{tabular}
\end{table}

\begin{figure*}[ht]
    \centering
    \includegraphics[width=1\textwidth]{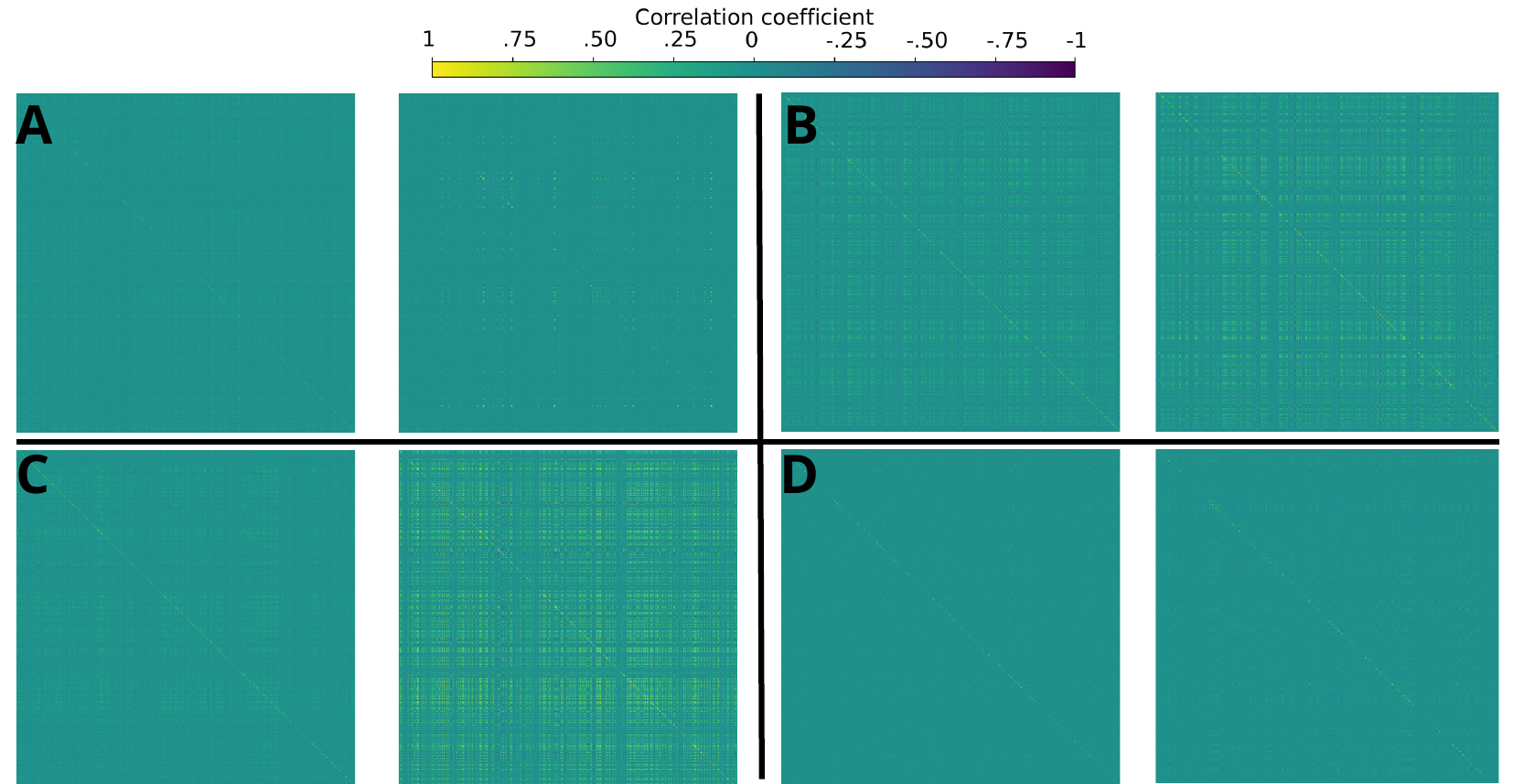}
    
    \caption{Pearson correlation matrices between spike trains of penultimate network layers, averaged over all tested samples for each tested model and dataset:
    \textbf{A} SewResnet18 MNIST; \textbf{B} SewResnet18 CIFAR10; \textbf{C} Spiking VGG 11 MNIST; \textbf{D} Spiking VGG 11 CIFAR10.  
    For each pair \textbf{A}-\textbf{D}, matrices are shown for correct prediction on clean samples (left) and noisy samples that led to incorrect prediction (right).}
    \label{fig:corr-matrices}
\end{figure*}

The internal dynamics of evaluated SNNs exhibit significant changes when subjected to adversarial noise conditions, particularly observable through the correlation patterns of spike trains within different layers. Examples of these spike train average correlation matrices are presented in Figure \ref{fig:corr-matrices}. For clarity and conciseness, we have chosen to display only the penultimate layers of each network; however, it is crucial to note that similar trends were observed across other layers as well, particularly within the fully connected layers, indicating a systemic response to input perturbations.

A primary observation from the presented plots in Figure \ref{fig:corr-matrices} is the pronounced increase in the number of neurons whose spike trains exhibit a positive correlation with other neurons within the same layer when the network processes noisy inputs. In the context of clean, unperturbed inputs, the correlation matrices often appear sparse, with fewer highly correlated spike trains, reflecting a more distributed and potentially independent neuronal activity. Conversely, under noisy input conditions, these matrices become visibly denser with positive correlations, suggesting a shift towards more synchronized or co-activated neuronal firing patterns. This heightened correlation implies that neurons within a layer are responding to the perturbed input in a more concerted manner, potentially as a compensatory mechanism or a symptom of disrupted information processing.

To quantitatively substantiate these visual observations, a detailed statistical analysis of these correlation matrix distributions was performed, with the results summarized in Table \ref{tab:corr_distribution_stats}. This analysis unequivocally demonstrates that noisy samples lead to a statistically significant increase in the total count of highly correlated spike trains. For instance, in the SewResnet18 model on the MNIST dataset, the number of values greater than 0.9 in the correlation matrix increased from 0 for clean samples to 15 for noisy samples. Similarly, for the Spiking VGG11 model on MNIST, this count surged from 0 to 2106. These figures provide quantitative evidence of the network's internal activity becoming more highly correlated under adversarial noise.

Furthermore, the statistical differences are reflected in changes in the kurtosis and skewness of the correlation value distributions. For noisy samples, both kurtosis and skewness tend to increase compared to clean samples. Kurtosis, which measures the "tailedness" of the distribution, indicates that the distribution of correlation values becomes more peaked around its mean and has heavier tails, meaning there are more extreme correlation values (both high positive and high negative, though the visual inspection of matrices suggests a prevalence of high positive correlations). An increased skewness, on the other hand, suggests that the distribution of correlation values becomes more asymmetric, with a longer tail extending towards higher positive correlation values. This shift in distribution characteristics provides a deeper statistical insight into the altered firing patterns, indicating a concentration of neuronal activity into more tightly coupled groups under the influence of noise. This phenomenon could be interpreted as the network attempting to maintain coherence or, conversely, as a breakdown of independent processing, leading to a more generalized and less discriminative response to the input.

\begin{table}[ht!]
\centering
\caption{Statistics of spike train correlation values in the penultimate network layers, averaged across all tested samples (clean samples with correct predictions, "Clean" column, and noisy ones with incorrect predictions, "Noise" column).}
\begin{tabular}{cccccccc}
\toprule
\multicolumn{2}{c}{Metric} & \multicolumn{2}{c}{MNIST} & \multicolumn{2}{c}{CIFAR-10} \\
\cmidrule(lr){3-4} \cmidrule(lr){5-6}
\multicolumn{2}{c}{} & Clean & Noise & Clean & Noise \\
\midrule
\multicolumn{6}{c}{\textbf{SewResnet18}} \\
\midrule
Kurtosis & & 8.20 & 161.82 & 6.36 & 7.18 \\
Skewness & & 2.58 & 10.33 & 2.05 & 2.13 \\
99-th percentile & & 0.16 & 0.13 & 0.30 & 0.43 \\
No. values $>$ 0.9 & & 0 & 15 & 0 & 53 \\
\addlinespace
\midrule
\multicolumn{6}{c}{\textbf{Spiking VGG11}} \\
\midrule
Kurtosis & & 5.08 & 4.01 & 43.95 & 52.02 \\
Skewness & & 1.72 & 1.74 & 1.34 & 1.65 \\
99-th percentile & & 0.22 & 0.68 & 0.12 & 0.16 \\
No. values $>$ 0.9 & & 0 & 2106 & 0 & 52 \\
\bottomrule
\label{tab:corr_distribution_stats}
\end{tabular}
\end{table}

\section{Discussion and Conclusion}

This study presents a systematic investigation into how neuron-level parameters - specifically the membrane time constant ($\tau$) and firing threshold voltage ($v_{\text{th}}$) - govern the internal dynamics and performance of SNNs. Our findings reveal that SNNs exhibit a well-defined ``operational manifold'' within their hyperparameter space. Operating within this manifold enables a favorable trade-off between task accuracy and energy efficiency (measured via spike count), while deviations result in network inactivity or excessive spiking, both detrimental to practical deployment.

The results demonstrate that both the membrane time constant and voltage threshold act as key control variables for the network's computational efficiency. Low $v_{\text{th}}$ values typically yield dense spiking and high accuracy but come at a significant energy cost. In contrast, excessively high $v_{\text{th}}$ suppresses neuronal firing altogether, leading to an effective shutdown of computational activity. Moderate thresholds, however, define a critical region where accuracy remains high while spike count - and thus energy consumption - is substantially reduced. This suggests the existence of an ``efficiency ridge'' within the hyperparameter space, a concept crucial for neuromorphic applications.

Architectural choices further modulate this behavior. CNNSNNs consistently outperformed MLPSNNs, especially on complex datasets like CIFAR10. However, this improved performance came at the cost of significantly higher spike counts, indicating a classic performance-efficiency trade-off. MLPSNNs, while less accurate, tended to exhibit more stable and energy-conservative behavior. These differences also suggest that optimal hyperparameter configurations are model- and dataset-specific, underscoring the necessity of systematic tuning.

Importantly, the two major lines of experimentation in this study - the grid-based exploration of hyperparameter space and the correlation-based analysis under adversarial noise - complement each other and converge on a key conclusion: 
\textbf{they jointly reveal the boundaries of the operational manifold.} 
While the grid search maps regions of effective and ineffective performance as functions of $\tau$ and $v_{\text{th}}$, the robustness experiments expose how models behave when pushed toward or beyond those boundaries under external perturbations. 
The rise in inter-neuronal correlation and the collapse of accurate prediction under noise are indicative of the network operating outside its stable manifold.

Beyond performance metrics, our work also assessed the robustness of SNNs to noisy inputs through a novel layer-wise analysis of spike train correlations. Results indicate a marked increase in inter-neuronal correlation under noisy conditions, suggesting that network responses become more synchronized and less discriminative when perturbed. This emergent behavior likely reflects a loss of representational diversity and may underpin the model’s susceptibility to misclassification. Such findings illuminate the internal collapse dynamics of SNNs under adversarial perturbations, a vital consideration for real-world deployment in safety-critical systems.

Our findings underscore the need for principled hyperparameter selection in SNNs, moving beyond trial-and-error or end-performance-centric AutoML approaches. A deeper understanding of internal dynamics not only enhances performance but also enables robust and energy-aware deployment in neuromorphic hardware environments.

Further research could focus on \textbf{fine-tuning thresholds around the ``knee-point''}\textemdash the region where accuracy remains high but spike activity (and consequently energy use) drops significantly.

\section*{Acknowledgments}
We gratefully acknowledge Polish high-performance computing infrastructure PLGrid (HPC Center: ACK Cyfronet AGH) for providing computer facilities and support within computational grant no. PLG/2024/017775.
\FloatBarrier
\bibliographystyle{IEEEtran}
\bibliography{bibliography}

\end{document}